\documentclass[10pt,twocolumn,letterpaper]{article}

\usepackage{cvpr} 
\usepackage{url}
\usepackage{graphicx}
\usepackage{amsmath}
\usepackage{amssymb}
\usepackage{booktabs}
\usepackage{caption}
\usepackage{algorithm}
\usepackage{algorithmic}
\usepackage[accsupp]{axessibility}

\usepackage[pagebackref,breaklinks,colorlinks]{hyperref}

\usepackage[capitalize]{cleveref}
\crefname{section}{Sec.}{Secs.}
\Crefname{section}{Section}{Sections}
\Crefname{table}{Table}{Tables}
\crefname{table}{Tab.}{Tabs.}

\def\cvprPaperID{1408} 
\def\confName{CVPR}
\def\confYear{2022}

\begin{document}

\title{DATA: Domain-Aware and Task-Aware Self-supervised Learning}

\author{
Qing Chang $^{1,3,4,5}$
\and
Junran Peng $^{2}$
\and
Lingxi Xie $^{2}$
\and
Jiajun Sun $^{1,3,4,5}$
\and
Haoran Yin $^{1,3,4,5}$ 
\and
Qi Tian $^{2}$
\and
Zhaoxiang Zhang\thanks{Corresponding author.} $^{1,3,4,5,6}$
\and
$^1$University of Chinese Academy of Sciences, $^2$Huawei Inc.\\
$^3$Institute of Automation, Chinese Academy of Sciences\\
$^4$National Laboratory of Pattern Recognition \\
$^5$Center for Research on Intelligent Perception and Computing\\
$^6$Centre for Artificial Intelligence and Robotics, HKISI\_CAS
\and
\texttt{changqing2020@ia.ac.cn}, \texttt{jrpeng4ever@126.com}\\
\texttt{198808xc@gmail.com},
\texttt{\{sunjiajun211,yinhaoran19\}@mails.ucas.ac.cn}\\
\texttt{tian.qi1@huawei.com},
\texttt{zhaoxiang.zhang@ia.ac.cn}
}

\maketitle

\begin{abstract}
    \vspace{-0.4cm}
   The paradigm of training models on massive data without label through self-supervised learning (SSL) and finetuning on many downstream tasks has become a trend recently. 
   However, due to the high training costs and the unconsciousness of downstream usages, most self-supervised learning methods lack the capability to correspond to the diversities of downstream scenarios, as there are various data domains, different vision tasks and latency constraints on models.
   Neural architecture search (NAS) is one universally acknowledged fashion to conquer the issues above, but applying NAS on SSL seems impossible as there is no label or metric provided for judging model selection.
   In this paper, we present DATA, a simple yet effective NAS approach specialized for SSL that provides \textbf{D}omain-\textbf{A}ware and \textbf{T}ask-\textbf{A}ware pre-training.
   Specifically, we (i) train a supernet which could be deemed as a set of millions of networks covering a wide range of model scales without any label, (ii) propose a flexible searching mechanism compatible with SSL that enables finding networks of different computation costs, for various downstream vision tasks and data domains without explicit metric provided. 
   Instantiated With MoCo v2, our method achieves promising results across a wide range of computation costs on downstream tasks, including image classification, object detection and semantic segmentation. DATA is orthogonal to most existing SSL methods and endows them the ability of customization on downstream needs.
   Extensive experiments on other SSL methods demonstrate the generalizability of the proposed method.
   Code is released at \url{https://github.com/GAIA-vision/GAIA-ssl}. 

\end{abstract}

\vspace{-0.5cm}
\section{Introduction}

\begin{figure}[]
    \centering
    \includegraphics[height=7cm]{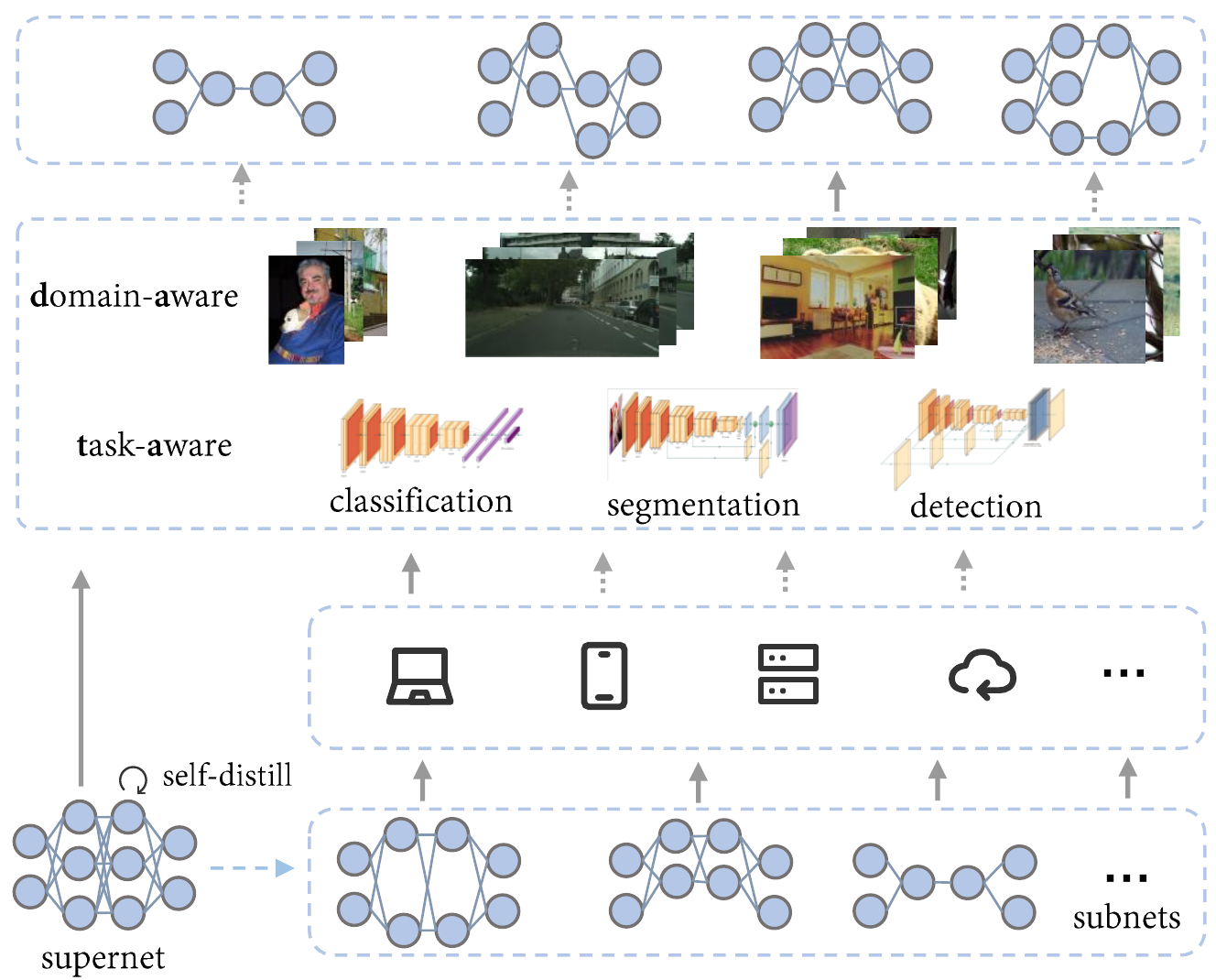}
    \caption{Illustration of how DATA works. We first build a supernet which is a set of many subnets, and train massive models simultaneously in the regime of self-supervised learning. Then, we propose a unsupervised searching method that enables domain-aware and task-aware model selection without any label. The mechanism enables self-supervised models to fit various scenarios including point,edge and cloud covering different vision tasks like image classification, object detection, and segmentation. Network architectures in figure are ploted by software PlotNeuralNet~\cite{haris_iqbal_2018_2526396}.}
    \label{fig:1_opening_pipeline}
\end{figure}

As is universally agreed that deep learning algorithms are data-hungry, how to leverage the exponentially growing unlabelled data from open-source has become a huge challenge. Self-supervised learning (SSL), which utilizes the inherent relationships of data to formulate supervision without manual annotation, has made remarkable progresses in both natural language processing (NLP)~\cite{floridi2020gpt,devlin2018bert,radford2018improving,radford2019language} and computer vision (CV)~\cite{he2020momentum, grill2020bootstrap, chen2020simple} area. 
Despite its great success, to unleash the true power of SSL requires gigantic scale of data and unimaginable training budgets. 
This brings about a side-effect that it is extremely expensive to train models of various architectures to cover the heterogeneous downstream needs, 
considering most scenarios are requesting for models of different scales  and different downstream vision tasks may desire different model architectures. 
It is usually believed that neural architecture search (NAS) is designed for solving the issues above. However, labels are so indispensable for existing NAS methods that it seems impossible to apply NAS in SSL, because there is no clue for model selection if no label or metric is provided.  
These reflections leave us two problems:

(1) \textbf{Is it possible to train networks of distinctive architectures simultaneously in SSL?} Gladly, previous methods~\cite{yu2020bignas, cai2019once, bu2021gaia,caron2021emerging} have proved that training a supernet that comprises of millions of weight-sharing subnets is possible in the regime of supervised learning. 
Thus the hardship only lies in how to prevent the co-training of different networks from diverging when there is no strong and stable supervision.   
Some recent studies~\cite{caron2021emerging, zheng2021ressl} interpret the process of SSL of siamese based methods as a form of self-distillation that the query-branch works as a student and the teacher-branch works as a teacher.
Thus if we stabilize the behaviour of the teacher, we could provide a relatively steady knowledge source for the heterogeneous students.
In this work, we build a supernet training mechanism for siamese based SSL that we fix the {\it key-branch} with the maximum architecture of supernet as a teacher and vary only the architectures of the {\it query-branch}. Experiments show that this ensures the efficiency of convergence and greatly improves the capability of feature representation of small subnets. 
More importantly, this design of training supernet in SSL brings us the answer to the critical question below.

\textbf{(2) How to judge the quality of a network if no label or metric is provided? } It is generally agreed that  {\it the bigger the better} works for deep neural networks when data is sufficient. Given a supernet that covers subnets of different sizes and the knowledge distillation behaviour of SSL, the distance between subnets and the maximum network naturally becomes a self-supervised metric for judging the quality of networks. This metric works well especially when there is a budget constraint for subnets. More discussions about this assumption are placed in Sec~\ref{sec:limitation}

We further extend our exploration to enable the searching process to be aware of the type of downstream tasks that different tasks adopt different types of features for measuring the distance of student and teacher. 
This greatly minimizes the gap transferring to downstream tasks while keeping the searching strategy plug-and-play. 

As shown in Figure~\ref{fig:1_opening_pipeline}, our approach enables training models of various sizes all in one go and searching appropriate models specialized for specific downstream tasks, computation constraints and data domains. This entire pipeline does not require any label for training or model selection. Instantiated with MoCo v2~\cite{chen2020improved} , we validate our contributions on evaluating our models on several standard of self-supervised benchmarks. 
We also combine our approach with other existing SSL methods~\cite{grill2020bootstrap, zheng2021ressl, wang2021dense} to demonstrate the generalizability.

\section{Related Work}

\subsection{Self-supervised learning}
Self-supervised learning has become the main paradigm of unsupervised learning. It aims at building a good pretext to learn fruitful feature representation from data itself. These pretexts can be mainly separated by two categories: reconstruction-based which concludes colorization~\cite{zhang2016colorful}, spatial jigsaw puzzles~\cite{wei2019iterative}, inpainting~\cite{yu2018generative} and discriminant-based which contains rotation predict~\cite{komodakis2018unsupervised}, instance-level contrastive~\cite{he2020momentum,chen2020simple,grill2020bootstrap}, and fine-grained contrastive~\cite{wang2021dense,xie2021propagate}. 

\paragraph{Contrastive learning.} For the first time, contrastive methods~\cite{he2020momentum,chen2020simple} 
makes self-supervised training become comparable with supervised counterpart. Their method mainly focus on pulling representations of different views of the same image (positive pairs) closer and pushing representations of different images (negative pairs) away in the same time. Further,~\cite{caron2020unsupervised,grill2020bootstrap} just use positive pairs to make network learn fruitful feature. These aforementioned methods pretext are sub-optimal in some extent for dense predict task (object detection, semantic segmentation, etc.). To relive this issue, fine-grained pretext~\cite{wang2021dense,xie2021propagate,liu2020self,bar2021detreg,henaff2021efficient} are proposed. Most of these methods already outperform supervised counterpart in some dense predict downstream tasks.

\subsection{Neural Architecture Search}

Neural architecture search aims at automating the architecture design process under certain constraints.~\cite{zoph2016neural,zoph2018learning} proposes to use reinforcement learning with the metric on proxy datasets as the reward for solving this problem. But due to the unaffordable cost, one-stage NAS~\cite{liu2018darts,cai2018proxylessnas,dong2019searching, akimoto2019adaptive, brock2017smash} are proposed, which train and search candidate architectures inside a single supernet. Though getting a specific architecture easily with those methods, we still need to train and search from scratch to get a new architecture once the constraints (such as latency, memory cost) changed. Further, researchers propose methods~\cite{yu2018slimmable,yu2020bignas,cai2019once,bu2021gaia,chen2021autoformer} to achieve training one supernet, which can contain a series of subnets that cover a wide range of scenarios. Their effective supervised training methodology inspired us the most. In~\cite{liu2020labels}, it firstly shows that network architectures performance on self-supervised tasks like rotation prediction is linearly correlated with the performance on the supervised task which inspired the design of searching mechanism in our method.

\section{Method}
\subsection{Preliminaries}
We formulate the process of common siamese based SSL as a process of dynamic knowledge distillation, which shares the same notion with~\cite{caron2021emerging}. To be convenient, we alias the {\it query-branch} as {\it student branch}, and the {\it key-branch} as {\it teacher branch} in our paper. 
Given $N$ unlabeled samples ${x_1,x_2,...,x_N}$, two views ($x_i^s$ and $x_i^t$) are obtained on each sample through composition of different augmentations $T$ and fed into a student $g(\cdot, {\theta^s})$ and a teacher network $g(\cdot, {\theta^t})$, parameterized by $\theta^s$ and $\theta^t$ respectively. In most cases, the teacher shares the exponential moving averaged (EMA) weights of student, namely, $\theta^t \leftarrow \lambda \theta^s + (1-\lambda)\theta^s$. 
We use $z_i^s = g(x^s, {\theta^s})$ and $z_i^t = g(x^t, {\theta^t})$ to denote the encoded features from student and teacher models, respectively. 
$H(z^s, z^t)$ is used to represent the similarity function.
Taking MoCo~\cite{he2020momentum} as an example, the InfoNCE loss~\cite{oord2018representation} (Eq.~\ref{eq1}) is adopted for training model:

\begin{equation}
\mathcal{L}_i=-\log \frac{\exp \left(H(g(x_i^s, {\theta^s}) \cdot g(x_i^t, \theta^t) / \tau\right)}{\sum_{j}^{N} \exp \left(H(g(x_i^s, {\theta^s}) \cdot g(x_j^t, {\theta^t})) / \tau\right)} \label{eq1}
\end{equation}
where $\tau$ is a temperature hyper-parameter~\cite{wu2018unsupervised}.

\subsection{Self-supervised Supernet Training}
No single model could perfectly match the needs of heterogeneous downstream applications, as there might be different latency constraints, data domains and task gaps. Thus we aim to train a great deal of models together instead of training a single one in the regime of SSL, and we hope they cover a wide range of model scale. 
We extend the definition of a network from $g(x, {\theta})$ to $g(x, {\theta}, \mathcal{A})$, with a new dimension meaning model architecture.  
Concretely, we build up a supernet $\Phi$ which contains numerous weight-sharing~\cite{yu2018slimmable} subnets $g^{(k)}$ of various architectures $\mathcal{A}^{(k)}$, formulated as:

\begin{equation}
\left\{\begin{array}{l}
\mathcal{A}=\left(\mathcal{A}^{(1)}, \ldots \mathcal{A}^{(k)}, \ldots \mathcal{A}^{(K)}\right) \\
\theta=\left(\theta^{(1)}, \ldots \theta^{(k)}, \ldots \theta^{(K)}\right)
\end{array}\right.\label{eq2}
\end{equation} where $K$ is the total number of subnets. 
Particularly, we mark the largest model in supernet $\Phi$ as $g(\cdot, \theta^{(K)}, \mathcal{A}^{(K)})$, because weights of all the subnets $g(\cdot, \theta^{(k)}, \mathcal{A}^{(k)})$ are completely included by it. During training, we fix the architecture of {\it teacher-branch} as the model-$K$ which means $\mathcal{A}^t = \mathcal{A}^{(K)}$, and vary the architecture of {\it student-branch} as $\mathcal{A}^s \in \mathcal{A}$. The weights of {\it teacher-branch} use the EMA version of the model-$K$, namely, $\theta^t \leftarrow \lambda \theta^{(K)} + (1-\lambda)\theta^{(K)}$. 

In each training iteration, we random sample two network architectures $\mathcal{A}^{(m)},\mathcal{A}^{(n)} $ from $\Phi$ together with the maximum one $\mathcal{A}^{(K)}$ to form a architecture set $\Omega = \{\mathcal{A}^{(m)}, \mathcal{A}^{(n)}, \mathcal{A}^{(K)}\}$. 
Following the conventional training regime in siamese-based SSL, we feed $x_i^t$ to {\it teacher-branch} and generate the embedded feature of teacher $z_i^t = g(x_i^t,\theta^t,\mathcal{A}^t)$, and
we feed $x_i^s$ to the models in {\it student-branch} from $\Omega$ to get student features $\mathcal{Z}_i^s = \{z_{i}^{s(m)},z_{i}^{s(n)},z_{i}^{s(K)}\}$ where $z_{i}^{s(m)} = g(x_i^s,\theta^{s(m)},\mathcal{A}^{(m)})$. With similarity measured by dot product, we apply InfoNCE loss on $\{(z_i^t, z_i^s) | z_i^s \in \mathcal{Z}_i^s\}$, back-propagate the gradients and update all the involved parameters.
The whole training process is shown as pseudo-code in Algorithm~\ref{algorithm1}

\begin{algorithm}

\caption{Self-Supervised Supernet Training}
\label{algorithm1}
\begin{algorithmic}[1]
\REQUIRE{Define supernet $\Phi$ with largest architecture $\mathcal{A}^{(K)}$. Choose the specific contrastive learning method to determine $criterion$}. Initialize the neural network $g(\cdot, \theta^s, \mathcal{A}^{(K)})$ and $g(\cdot, \theta^t, \mathcal{A}^{(K)})$

\FOR{$i = 1,...,T_{iters}$.}
\STATE{Get the min-batch of data $x_i$.}
\STATE{Get two views of $x_i$. $x_i^s$, $x_i^t$.}
\STATE{$optimizer.zero\_grad()$.}
\STATE{$loss$ initialized with 0.}
\STATE{ $z_{t}=g(x_i^t,\theta^t,\mathcal{A}^{(K)})$}. 

\STATE{Sample two model architectures $\mathcal{A}^{(m)},\mathcal{A}^{(n)}$ from $\Phi$ to construct set $\Omega=\{\mathcal{A}^{(m)},\mathcal{A}^{(n)},\mathcal{A}^{(K)}$\}}.

\FOR{$\mathcal{A}^{(k)}$  in $\Omega$}
\STATE{$loss += criterion(z_{t}, g(x_i^s,\theta^{s(k)},\mathcal{A}^{(k)}))$.}
\ENDFOR

\STATE{$loss.backward()$.}
\STATE{$optimizer.step()$. }
\STATE{$\theta^t \leftarrow \lambda \theta^s + (1-\lambda)\theta^s$.}
\ENDFOR

\end{algorithmic}
\end{algorithm}

\paragraph{Model space of supernet.} We choose the popular ResNet~\cite{he2016deep} as the basic architecture in our work. Depth\footnote{Number of $bottleneck$ blocks in each stage.} and width\footnote{Number of channels of $3\times3$ convolutions in each stage.} are adopted as factors to formulate the model space. Sharing the same notion with~\cite{he2016deep}, the output feature maps of each stage are denoted as $(C1, C2, C3, C4, C5)$ for future use. As shown in Table~\ref{tab:3_model_space}, the depth of stage start from (2, 2, 5, 2) to (4, 6, 29, 4) with a step of (1, 2, 2, 1), and the width of $stem$ and each stage start from (32, 48, 96, 192, 384) to (64, 80, 160, 320, 640) with a step of (16, 16, 32, 64, 128). 

\setlength\tabcolsep{5pt}
\begin{table}[!ht]
    \centering
    \begin{tabular}{l |c c| c c}
    \bottomrule
      layer name  & $W_{range}$ & $W_{step}$  & $D_{range}$ & $D_{step}$ \\ \hline
       $stem$ & [32, 64] & 16 & - & - \\ 
        $stage1$ & [48, 80] & 16 & [2, 4] & 1\\ 
        $stage2$ & [96, 160] & 32 & [2, 6] & 2\\
        $stage3$ & [192, 320] & 64 & [5, 29] & 2\\
        $stage4$ & [384, 640] & 128 & [2, 4] & 1 \\ \bottomrule
    \end{tabular} 
    \caption{Model space of supernet. Width and depth of each stage are sampled in range with certain step. }\label{tab:3_model_space} 
\end{table}

\begin{figure*}[]
    \centering
    \includegraphics[height=8cm]{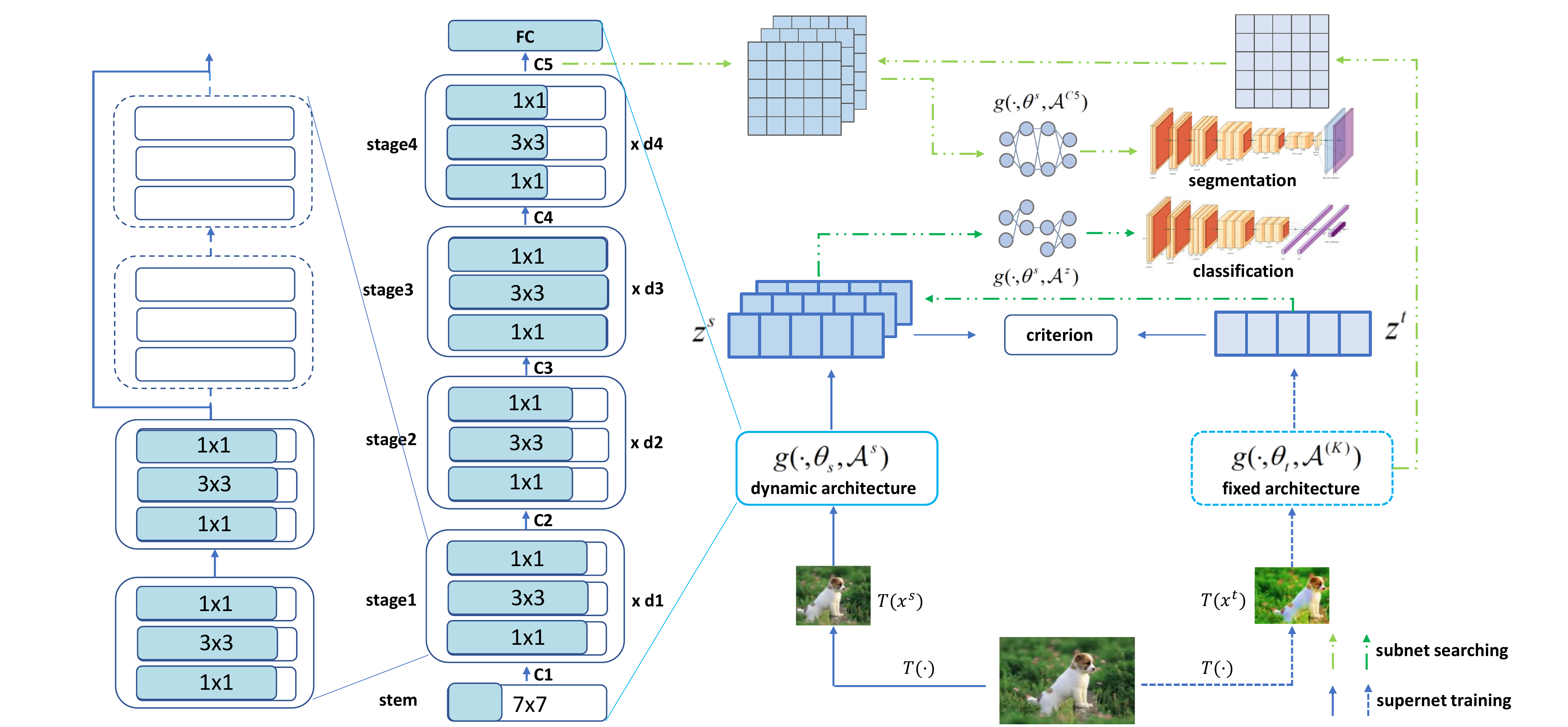}
    \caption{Pipeline of our method. It contains two stages. 
    In the first stage, we fix the architecture of the {\it key-branch} as a steady teacher, and vary the architectures of the {\it query-branch} for supernet training. 
    In the second stage, we propose a domain-aware and task-aware self-supervised metric for subnet searching, based on the similarity of task-specific features extracted from target dataset between subnets and model of the {\it key-branch}. d\{1,2,3,4\} : number of $bottleneck$ blocks in this stage. 
    }
    \label{fig:3_pipeline_of_our_method}
\end{figure*}

\subsection{Self-supervised Model Selection}
This part reveals the core motivation behind our adoption of supernet in the regime of SSL. Beyond the value of training massive models simultaneously, 
it presents a feasible metric for self-supervised NAS and provides abundant architecture candidates for searching. As is universally acknowledged that {\it the bigger the better} works mostly for deep neural networks, the distance between subnets and the largest model naturally becomes a metric for model selection.

\paragraph{Domain-awareness.}
This mechanism enables us to apply NAS on the downstream data during self-supervised learning, which shrinks the domain gap.
Specifically, given a constraint of computation budget $C$, 
we feed the downstream data $\mathcal{D}$ into {\it teacher branch} and obtain the exemplar feature representations 
$z_i^t=g(x_i,\theta^t,\mathcal{A}^{(K)}) $ for  $x_i \in \mathcal{D}$.
Then we randomly sample subnets under the constraint of budget and collect feature representations
$z_i^{s(k)}=g(x_i,\theta^{s(k)},\mathcal{A}^{(k)})$ for
each model on the downstream data. We denote the similarity function by $H'$ and the conquering model is selected to maximize the similarity on the entire downstream dataset:
\begin{equation}
\max_{k} \sum_i^{|D|}{H'(z_i^t, z_i^{s(k)})}
\label{feature distance}
\end{equation}
The whole process does not involve any fully-supervised metric such as accuracy or precision.

\paragraph{Task-aware metrics.}
The gaps between different vision tasks are always unignorable. 
To enable our pre-trained supernet to serve various downstream vision tasks, the architecture search has to be conducted under task-aware metrics. 
Model selected under a single self-supervised metric is not suitable for different tasks.
For instance, classifier of image classification task mostly handles the $avg$-$pooled$ feature that focuses on the global information of image, while in object detection, equipped with  FPN~\cite{lin2017feature}, detectors use the multi-stage features of backbone for inference. 
These task-specific types of features matter most for the specific downstream task. 
Hence we search for different models by measuring the distance of the features directly used in head of downstream task. We call this task-aware metrics. The pipeline of our method is demonstrated in Figure~\ref{fig:3_pipeline_of_our_method}.

For task of image classification, we direct use the feature of $z$. For object detection, we adopt feature of C5 for Faster-RCNN-C4 and features from C2-C5 for Faster-RCNN-FPN. And for task of semantic segmentation, we adopt features from C4-C5.   
The influence of task-aware metric is detailed in Table~\ref{tab:5_ablation_with_task-aware_metrics}.

\paragraph{Similarity of relative relations.} During the task-aware searching, the size of the feature map of subnets might be inconsistent with it of teacher. 
Thus we turn to utilize the relative relations of features to evaluate the similarity of two feature maps. 
We use $M \in \mathbb{R}^{C' \times HW}$ and $M'\in \mathbb{R}^{C \times HW}$ to represent features from student and teacher respectively, where $H$,$W$ denote the height and width, $C$ and $C'$ represent number of channels, and $\{m_1,m_2,...,m_{HW}\}$ means the feature in each column of matrix $M$. 
We define $r_{ij}$ to express the vector relative relation of vector $m_j$ and $m_i$ as in Eq.~\ref{relative relationship}, and the similarity of relative relation $R$ on feature maps can be formulated in Eq.~\ref{relative relationship similarity}.
We combine this with the task-aware metric, and conduct model selection in Eq.~\ref{feature distance}.

\begin{equation}
r_{ij} = -\log \frac{\exp \left(m_i \cdot m_j / \tau\right)}{\sum_k^{HW}{\left(m_i \cdot m_k / \tau\right)}}
\label{relative relationship}
\end{equation}

\begin{equation}
R(M,M') =\sum_i^{HW}{\sum_j^{HW}-r'_{ij}\log{r_{ij}}}
\label{relative relationship similarity}
\end{equation}

\section{Experiments}

\subsection{Experiments Setup}
\noindent \textbf{Datasets.} We instantiate our method with the popular MoCo v2~\cite{chen2020improved}  and train on ImageNet~\cite{deng2009imagenet} which has $\sim$1.28 million images in 1000 categories.
For the purpose of verifying transferability of models we search on downstream tasks, we experiment on ImageNet~\cite{deng2009imagenet} semi-supervised classification, COCO~\cite{lin2014microsoft} instance segmentation, PASCAL VOC~\cite{everingham2010pascal} object detection, and Cityscapes~\cite{cordts2016cityscapes} semantic segmentation. In the ablation study, to validate the generalizability of our method, we combine our method with other self-supervised learning methods ~\cite{grill2020bootstrap,zheng2021ressl,wang2021dense} and train them on ImageNet-$10\%$.

\paragraph{Training details.}  When supernet trained on Imagenet, we use SGD as the optimizer. The SGD weight decay is 0.000075. We train 200 epochs using a batch size of 1024 on 16 GPUs and an initial learning rate of 0.12. For supernet trained on ImageNet-$10\%$, we all follow their official default settings.

\paragraph{Searching details.}
In order to verify that our method can effectively deal with various scenarios, we compute FLOPs for all of our subnets according to 224x224 input resolution and divide them into seven groups at 1 GFLOPs interval from 1 GFLOPs $\sim$ 8 GFLOPs. We randomly sample one hundred subnets in different groups and search the network in target dataset according to the task-aware metric to find the best one in every group. When verifying our method on various downstream tasks, we will display our experiment results of each group. 

\subsection{Ranking Correlation}
Firstly, we verify that the similarity of the task-specific features between pre-trained subnets and the largest network can be a reliable indicator to assess the performance of these subnets. We sample fifty subnets uniformly and experiment on the above ImageNet-$1\%$ (IN-$1\%$) semi-supervised classification, VOC object detection, COCO instance segmentation and Cityscapes semantic segmentation. We compute the Spearman~\cite{spearman1904proof} correlation between the similarity ranking and their final performance ranking. The results are shown in Table~\ref{tab:4_ranking_correlation}.  
For the low ranking correlation of semantic segmentation, we infer the reason is that the data distribution of the Cityscapes~\cite{cordts2016cityscapes} dataset is largely different from the ImageNet~\cite{deng2009imagenet} dataset used in the self-supervised training stage. 
\setlength\tabcolsep{4.5 pt}
\begin{table}[!ht]
    \small
    \centering
    \begin{tabular}{l l c c}
    \bottomrule
        Architecture & Dataset & Feature & Correlation  \\ 
        \hline
        ResNet-FC~\cite{he2016deep}&IN-$1\%$\cite{deng2009imagenet}  & $z$  & 0.90 \\ 
        FasterRCNN-C4~\cite{ren2015faster}  & VOC~\cite{everingham2010pascal} & C5 & 0.84\\ 
        MaskRCNN-FPN~\cite{He_2017_ICCV}& COCO~\cite{lin2014microsoft} & C2-C5&
        0.86\\
        FCN~\cite{long2015fully} & Cityscapes~\cite{cordts2016cityscapes}  & C4-C5 & 0.63\\
    \bottomrule
    \end{tabular} 
    \caption{Ranking correlation. Architecture : The specific algorithm architecture adopted when transferring to downstream tasks. Feature : The feature utilized for model selection. 
    IN-$1\%$ : ImageNet-$1\%$.}
    \label{tab:4_ranking_correlation} 
\end{table}

\subsection{Results on Various Downstream Tasks}

\noindent \textbf{Linear evaluation on ImageNet.} We train supervised linear classifiers (a fully-connected layer with softmax) on frozen features after BN statics calibration for all of networks we search, following the procedure described in~\cite{yu2019universally}  (detailed in supplementary materials). We report 1-crop, top-1 classification accuracy on the ImageNet validation set. The result is shown in Figure~\ref{fig:4_linear_evaluation_on_imagenet} and the network architectures we search are described in 
appendix. We notice .The top-1 accuracy of linear evaluation of the model we search in Group 3G-4G is $68.5\%$ which outperforms ResNet50 (3.8G, $67.5\%$) by $1\%$.

\setlength\tabcolsep{3.5pt}
\begin{figure}[!t]
    \centering
    \includegraphics[width=8.5cm]{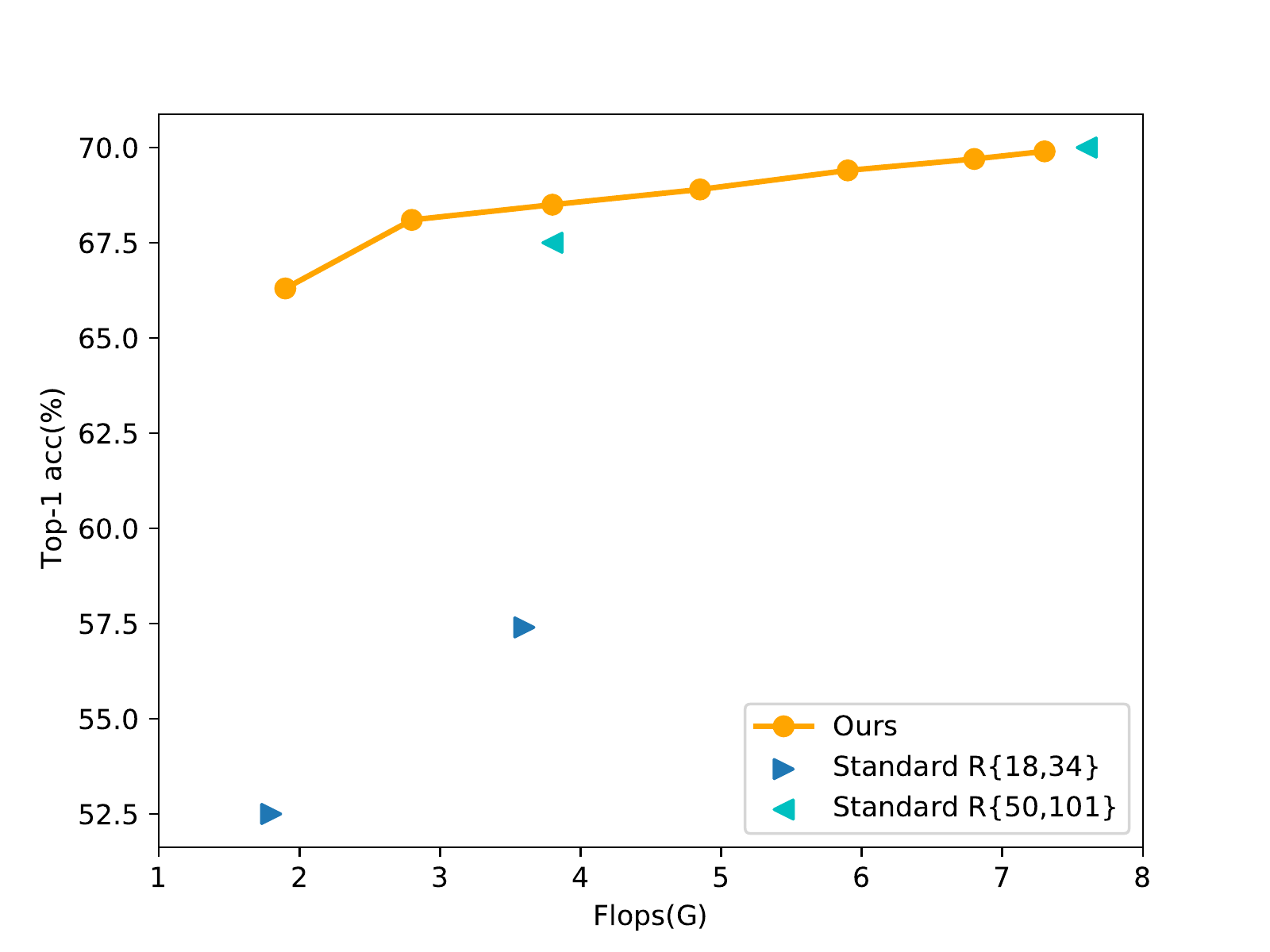}
    \caption{Linear evaluation on ImageNet. All of the classifiers are trained on the ImageNet train set with 100 epochs and evaluated on the ImageNet validation set. We compare it with standard ResNet architectures, which are all trained with MoCo v2 on ImageNet for 200 epochs. }
    \label{fig:4_linear_evaluation_on_imagenet}
\end{figure}

\setlength\tabcolsep{4pt}
\makeatother
\begin{table*}[!ht]
    \centering
    \begin{tabular}{l |c c c c| c c c| c c c}
    \bottomrule
        Model  & FLOPs &Params & Depth & Width & $\mathrm{AP^b}$ & $\mathrm{AP^b_{50}}$ & $\mathrm{AP^b_{75}}$ & $\mathrm{AP^m}$ &$\mathrm{AP^m_{50}}$ &$\mathrm{AP^m_{75}}$ \\ \hline

        R50*~\cite{chen2020improved} & 3.8G & 25.5M& [3, 4, 6, 3]&[64, 64, 128, 256, 512] & 38.7  &59.2 & 42.3 & 35.5 & 56.2 & 37.9\\ 
        R50$\dag$~\cite{wu2019detectron2} & 3.8G & 25.5M& [3, 4, 6, 3]&[64, 64, 128, 256, 512] & 38.6  &59.5 & 42.1 & 35.2 & 56.3 & 37.5\\ \hline
       
         Group  & FLOPs &Params & Depth & Width & $\mathrm{AP^b}$ & $\mathrm{AP^b_{50}}$ & $\mathrm{AP^b_{75}}$ & $\mathrm{AP^m}$ &$\mathrm{AP^m_{50}}$ &$\mathrm{AP^m_{75}}$ \\ \hline
        1G$\sim$2G & 1.8G & 13.6M &[2, 2, 5, 2] &[32, 48, 96, 192, 512] & 36.2 & 56.6 & 39.5 & 33.4 & 53.9 & 35.4   \\ 
        2G$\sim$3G & 2.7G & 14.7M & [2, 2, 13, 2] & [48, 48, 96, 192, 384]& 38.3&58.4 & 41.8 & 34.8 & 55.2 & 37.4 \\ 
        3G$\sim$4G & 3.7G & 25.7M & [3, 2, 17, 3] & [32, 48, 96, 192, 512] & \textbf{39.9}& \textbf{60.2} & \textbf{43.5} & \textbf{36.0} & \textbf{57.1} & \textbf{38.6}\\ 
        4G$\sim$5G & 4.2G & 33.1M & [2, 2, 25, 2] & [64, 64, 128, 192, 384] & 40.4 & 60.9 & 44.3 & 36.5 & 58.0 & 39.3  \\
        5G$\sim$6G & 5.9G & 43.4M & [4, 6, 21, 4] & [32, 64, 96, 192, 640] & 41.2 & 61.9 & 45.2 & 37.2 & 58.5 & 39.9 \\ 
        6G$\sim$7G & 6.6G & 40.2M & [3, 6, 27, 3] & [64, 80, 96, 192, 640]  & 41.5 & 62.1 & 45.1 & 37.3& 58.9 & 40.1\\ 
        \bottomrule
    \end{tabular} 
    \caption{Results of object detection and instance segmentation on COCO. 
    * : Results of model pretrained through MoCo v2. 
    $\dag$ :   Results of model pre-trained on ImageNet, fine-tuned on $train2017$ following 1x schedule and evaluated on $val2017$. 
    }\label{tab:4_instance_segmentation_fine_tuned_on_coco} 

\end{table*}
We find that relative small models benefit most from the training strategy. But the searched architecture in 7G$\sim$8G group performs slightly worse than R101. We infer that when one model architecture outperforms the largest one, its output features are also far away from the largest teacher.

\paragraph{Semi-supervised classification on ImageNet.} Next, following~\cite{chen2020simple}, we evaluate performance of the architectures we search on ImageNet $val$ set after fine-tuning on $1\%$ and $10\%$ subset of ImageNet $train$ set with annotations. For the convenience of description, we use ImageNet-$1\%$ and ImageNet-$10\%$ to represent the two subsets. The training procedure is detailed in supplementary materials. The results of Top-1 on the $val$ set are reported in Table~\ref{tab:4_semi_supervised_learning_on_ImageNet}. We obtain $+7.6\%$ boost when fine-tuned on ImageNet-$1\%$ and gain $+1.7\%$ on ImageNet-$10\%$. For ImageNet-$1\%$ semi-supervised setting, even the model selected from 1G$\sim$2G group outperforms the ResNet50 baseline by a large margin.

\setlength\tabcolsep{0.5pt}
\begin{table}[!t]
    \centering
    \begin{tabular}{l c c c c c}
    \bottomrule
        Model  &\quad Params & \quad \quad \quad \quad Top-1 & &\quad \quad \quad Top-5\\
        & & $1\%$ & $10\%$ & $1\%$ & $10\%$  \\ \hline
        R50*~\cite{chen2020improved} &\quad 25.5M & 39.8 & 61.8 & 68.3 & 85.1 \\ 
        \hline
        Group &\quad Params & $1\%$ & $10\%$ & $1\%$ & $10\%$  \\ \hline
        1G$\sim$2G & \quad14.7M &45.8 & 61.9 &73.7 &84.9  \\ 
        2G$\sim$3G &\quad 19.5M & 46.5 & 63.0&74.7&85.8 \\
        3G$\sim$4G &\quad 33.5M & \textbf{47.5} & \textbf{63.4} & \textbf{75.1} & \textbf{85.7}\\
        4G$\sim$5G &\quad 37.0M & 47.8 & 64.3 &75.3&86.4\\
        5G$\sim$6G & \quad 43.4M & 49.0 & 65.1 &76.2&86.7\\ 
        6G$\sim$7G & \quad 45.4M & 49.3 & 65.4 &76.3&87.0\\ 
        7G$\sim$8G &\quad 45.2M & 49.5 & 65.6 &76.5&86.9\\ \bottomrule
    \end{tabular} 
    \caption{
    Results of semi-supervised classification on the $1\%$ and $10\%$ portion ImageNet.
    * : Results of models pre-trained through MoCo v2. 
    $\dag$ : Results of models pre-trained on ImageNet.
    All of the models are fine-tuned on the corresponding subset for 20 epochs and evaluated on the ImageNet validation set. }\label{tab:4_semi_supervised_learning_on_ImageNet} 
\end{table} 

\paragraph{COCO instance segmentation.}
For COCO instance segmentation, we follow the common setting~\cite{he2020momentum} that fine-tune a Mask R-CNN detector (FPN) on COCO $train2017$ split ($\sim$118k images) for all the architectures of our search with standard $1\times$ schedule and evaluating on COCO $val2017$ split. The results are shown in Table~\ref{tab:4_instance_segmentation_fine_tuned_on_coco}. The pre-train architecture under 4 GFLOPs of our search outperforms the standard ResNet50 by $1.2 AP$ in detection, and the promotion is $0.6 AP$ for segmentation. 

\setlength\tabcolsep{4.5 pt}
\begin{table}[!ht]
    \centering
    \begin{tabular}{l| c c| c c c}
    \bottomrule
        Model  & FLOPs & Params & $\mathrm{AP}$ & $\mathrm{AP_{50}}$ & $\mathrm{AP_{75}}$  \\ \hline
        R50$\dag$\cite{chen2020improved}  &3.8G & 25.5M &53.5  &81.3  & 58.8  \\
        R50*\cite{chen2020improved} &3.8G & 25.5M & 57.2 & 82.4 & 63.7 \\
         \hline
        Group & FLOPs & Params & $AP^b$ & $AP^b_{50}$ & $AP^b_{75}$ \\ \hline
        1G$\sim$2G &1.8G &12.2M & 50.9 & 79.8 & 55.2  \\ 
        2G$\sim$3G & 2.8G & 25.0M & 58.0 & 82.7 & 64.5\\ 
        3G$\sim$4G & 3.7G & 25.6M & \textbf{58.5} & \textbf{83.0} & \textbf{65.0}  \\ 
        4G$\sim$5G &4.2G &33.1M & 59.1&83.2&65.3\\
        5G$\sim$6G &5.9G & 43.4M & 60.5 & 83.9 & 67.1  \\ 
        6G$\sim$7G &6.9G & 47.4M & 60.4 & 83.5 & 67.0 \\ 
        7G$\sim$8G &7.3G & 45.3M & 60.4&83.6&67.3 \\ \bottomrule
    \end{tabular} 
    \caption{Results of object detection on PASCAL VOC. 
    * : Model pretrained by MoCo v2.
    $\dag$ : Results of models pre-trained on ImageNet.
    }
    \label{tab:4_pascal_voc_object_detection} 
\end{table} 

\paragraph{Object detection on PASCAL VOC.}
When transferring to VOC~\cite{everingham2010pascal} object detection, following the~\cite{chen2020improved}: a Faster R-CNN detector~\cite{ren2015faster} (C4-backbone) is adopted. And it is trained on VOC $trainval$07+12 set for 24K iterations and evaluated on the VOC $test$2007 set. The results are presented in Table~\ref{tab:4_pascal_voc_object_detection}. With our framework, we can provide a model whose FLOPs and Params are similar to standard ResNet50, outperforms the R50 baseline by 1.2 AP.

\setlength\tabcolsep{3.5pt}
\begin{figure}[]
    \centering
    \includegraphics[width=8.5cm]{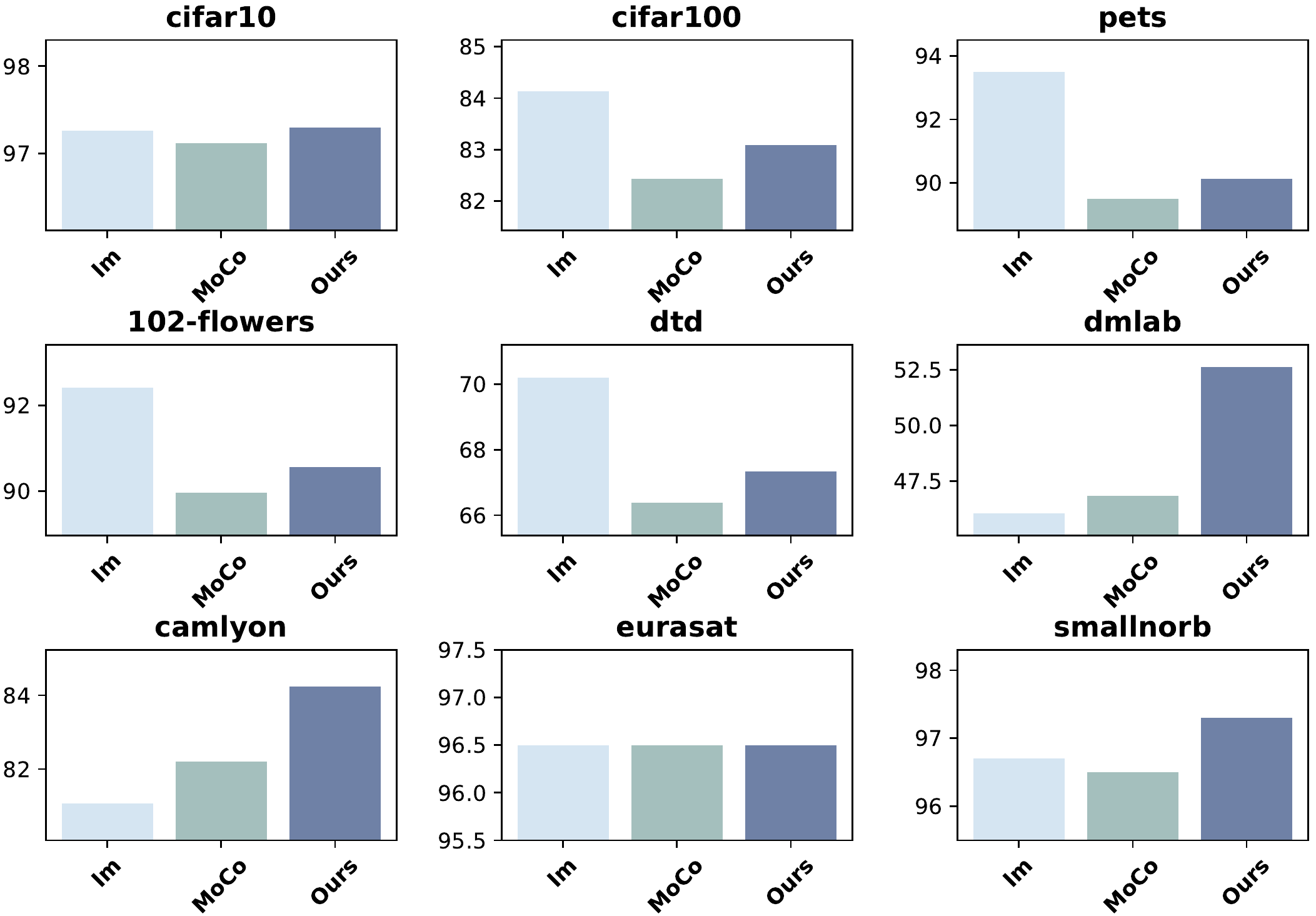}
    \caption{
    Transferablity to other data domains in image classification task. 
    IN : Model pre-trained on ImageNet.
    MoCo : Model pre-trained through MoCo v2.}
    \label{tab:4_transfer_to_other_classification_tasks}
\end{figure}

\setlength\tabcolsep{4.5 pt}
\begin{table}[!ht]
    \centering
    \begin{tabular}{l | c c| c }
    \bottomrule
        Model  & Depth & Width & mIoU  \\ \hline
         R50$\dag$  &[3, 4, 6, 3] & [64, 64, 128, 256, 512] &  75.5 \\ 
        R50*~\cite{chen2020improved} &[3, 4, 6, 3] & [64, 64, 128, 256, 512] &  76.4\\  \hline
        Group & Depth & Width & mIoU \\ \hline

        1G$\sim$2G &[2, 2, 5, 2] &[48, 48, 96, 192, 512] & 72.7  \\ 
        2G$\sim$3G &[3, 2, 9, 2] & [48, 48, 96, 192, 512] & 75.2 \\ 
        3G$\sim$4G &[2, 2, 17, 3] & [32, 48, 96, 192, 384] &\textbf{77.4} \\ 
        4G$\sim$5G & [2, 6, 19, 3]&[32, 48, 96, 192, 640] &77.0 \\
        5G$\sim$6G &[2, 4, 25, 3]  & [64, 64, 128, 192, 640] &78.1 \\ 
        6G$\sim$7G &[4, 6, 23, 4] & [32, 64, 128, 192, 640] & 77.6\\ 
        7G$\sim$8G &[4, 6, 21, 4] &[32, 64, 160, 192, 640] & 78.2 \\ \bottomrule
    \end{tabular} 
    \caption{
    Results of semantic segmentation on Cityscapes. 
    $\dag$ : Models pre-trained on ImageNet. 
    * : Models pre-trained through MoCo v2.
    }\label{tab:4_cityscapes_semantic_segmentation} 
\end{table} 

\paragraph{Cityscapes semantic segmentation.} Cityscapes~\cite{cordts2016cityscapes} is a widely used benchmark for semantic segmentation. Following~\cite{wang2021dense}, we fine-tune the backbone of our search in the FCN~\cite{long2015fully} form on $train$  set (2975 images) for 40k iterations with batch size 16 and test on $val$ set (500 images). The results are reported in Table~\ref{tab:4_cityscapes_semantic_segmentation}. Although the correlation of our ranking strategy is 
fair, as shown in Table~\ref{tab:4_ranking_correlation}, we can still find the network outperforms the baseline with our effective training by 1 mIoU.

\paragraph{Transfer to other classification tasks.} 
Furthermore, we evaluate our framework on more diverse classification datasets in VTAB~\cite{zhai2019large} (detailed in supplementary materials). We only find the most similar pre-train architecture in 3G$\sim$4G FLOPs group for the comparison with standard ResNet50. We perform fine-tuning on these datasets and report the results in Figure~\ref{tab:4_transfer_to_other_classification_tasks}. The training details are detailed in supplementary materials. Compared to our MoCo v2~\cite{he2020momentum} baseline, we can get consistent improvement on these datasets. 
Although MoCo v2 is at a disadvantage in classification compared to supervised pre-training, thanks to the advantages of our training strategy and model customization, it can exceed the performance of supervised pre-training on some datasets. As the results show, the performance on eurasat~\cite{helber2019eurosat} dataset 
attracts our attention. This dataset is a land cover classification dataset, and we find its data distribution is completely different from ImageNet~\cite{deng2009imagenet}. Hence, when transferring to this dataset, these feature representations are of no use.

\section{Ablation Study}\label{ablation study}
\noindent \textbf {Combining with contrastive learning methods.}
Our proposed pipeline is orthogonal to most contrastive learning methods. 
The reason for choosing MoCo v2 as baseline lies in its comprehensive performance on various downstream tasks. 
To demonstrate the generalizability of DATA, we instantiated with three classic contrastive self-supervised learning methods, BYOL~\cite{grill2020bootstrap}, ReSSL~\cite{zheng2021ressl} and DenseCL~\cite{wang2021dense}.
Due to the huge computation cost for training self-supervised models, we conduct experiments of these methods only on ImageNet-$10\%$. 
Since ~\cite{grill2020bootstrap, zheng2021ressl} are designed for classification task originally and~\cite{wang2021dense} is mainly designed for the dense-prediction task, we report comparison results on area of their expertise. 
As shown in Table~\ref{tab:5_ablation_on_contrastive_learning_methods}, consistent improvement could be observed.

\setlength\tabcolsep{8.5 pt}
\begin{table}[!ht]
    \centering
    \begin{tabular}{l c c}
    \bottomrule
        Method  & Task & performance  \\ \hline
        BYOL (repro) & Cls & 23.7 \\ Ours+BYOL  & Cls & \textbf{24.8}\\ \hline
        ReSSL (repro) & Cls & 23.7 \\ Ours+ReSSL & Cls & \textbf{26.7} \\ \hline
        DenseCL (repro) & Det & 49.1\\
        Ours+DenseCL & Det & \textbf{50.2}\\
    \bottomrule
    \end{tabular} 
    \caption{
    Ablation study on combining DATA with other contrastive learning methods. 
    Cls : Results of semi-supervised classification on the ImageNet-$1\%$ dataset. The top-1 accuracy adopted as metric. Det : Results of object detection on PASCAL VOC. AP is adopted as metric.  }
    \label{tab:5_ablation_on_contrastive_learning_methods} 
\end{table} 

\paragraph{Ablation with teacher architecture.}

In this ablation study, we explore the impact of choice on teacher architecture. Instead of fixing the architecture of {\it teacher branch} to the largest network, we compare with the setting using the same architecture as the sampled subnet in {\it student branch}, namely, $z_{i}^{t(k)} = g(x_i^t,\theta^{t(k)},\mathcal{A}^{(k)})$.
We base the experiment on MoCo v2 and train supernets on ImageNet-$10\%$.
Next, we extract the standard ResNet50 from supernet and evaluate it on the {\it val} set following the setting of linear classification.
As reported in Table~\ref{tab:5_ablation_with_feature_alignment}, we see that fixing the architecture of {\it teacher branch} is crucial, that the top-1 accuracy reaches $42.4\%$ under linear evaluation protocol, $3.4\%$ higher than the unfixed setting. This convey a message that a stable teacher is important for self-supervised supernet training. 
We also observe that this result outperform the vanilla MoCo v2 by $0.9\%$, which means supernet distillation is helpful over self-distillation.

\setlength\tabcolsep{5.5 pt}
\begin{table}[!ht]
    \centering
    \begin{tabular}{l c c c c}
    \bottomrule
    Method  & Fixed teacher arch & Top-1 & Top-5 \\ \hline
    MoCo v2 & &$41.5\%$ & $66.6\%$ \\ \hline
    Ours &  & $39.0\%$ & $64.2\%$ \\

    Ours & \checkmark  & \textbf{$42.4\%$} & \textbf{$67.2\%$} \\
    \bottomrule
    \end{tabular} 
    \caption{Ablation on feature alignment. This table reports  top-\{1,5\} accuracy of linear evaluation with 200 epochs on ImageNet-$10\%$.}\label{tab:5_ablation_with_feature_alignment} 
    
\end{table}

\textbf{Ablation with task-aware metrics.}
Here we explore the effectiveness of task-aware metrics for model selection. 
For each downstream task, we select models according to metrics based on $z$, C5 and C2-C5 respectively. 

As reported in Table~\ref{tab:5_ablation_with_task-aware_metrics}, task-aware metrics yield $0.4\%$, $0.7\%$ and $0.7\%$ improvements when task and metric are matched.

\setlength\tabcolsep{3 pt}
\begin{table}[!ht]
    \small
    \centering
    \begin{tabular}{l c c | c c c}
    \bottomrule
        Dataset &Task & Architecture & Feature & Performance \\ \hline
         &  &  &$z$ & 47.5\\
        IN-$1\%$~\cite{deng2009imagenet} & Semi-cls & ResNet-FC & C5& 47.3  \\ 
         &  & &C2-C5 & 47.1 \\ \hline
         &  & &  $z$ &57.8 \\
        VOC~\cite{everingham2010pascal}& Det & FasterRCNN-C4& C5 & 58.5  \\ 
        & &   &  C2-C5 & 58.1 \\ \hline
       &  &  &$z$ &39.2  \\
         COCO~\cite{lin2014microsoft} & Det &MaskRCNN-FPN & C5  &39.3 \\ 
         &  & & C2-C5 & 39.9\\        
   
    \bottomrule
    \end{tabular} 
    \caption{Ablation with task-aware metrics. IN-$1\%$ : ImageNet-$1\%$. Semi-cls : Semi-supervised classification, using top-1 accuracy as metric. Det : Object detection, using AP@IoU as metric. All models share the similar computation budget with ResNet50.}
    \label{tab:5_ablation_with_task-aware_metrics} 
\end{table} 

\paragraph{Ablation with domain-awareness.} 
We also explore the influence of domain-awareness. 

Specifically, we compare our searched models above with models searched through different datasets. Results are reported in Table~\ref{tab:5_ablation_with_task-aware_metrics}. Note that these models are all in 3G$\sim$4G group.

We find that searching by ImageNet seems to be an acceptable indicator for the performance of object detection on COCO. 
While in task of segmentation, searching without domain-awareness severely degrades the correlation($0.63 \rightarrow 0.23$) and mIoU($77.4 \rightarrow 76.2$). 

\setlength\tabcolsep{2.5 pt}
\begin{table}[!ht]
    \centering
    \begin{tabular}{l c c c c c}
    \bottomrule
        Task  & Target & Source &  Correlation & Performance \\ \hline
        Det & COCO &ImageNet & 0.82 & 39.7\\
        Det & COCO &COCO & 0.86 &39.9\\ \hline
        Seg & Cityscapes & ImageNet& 0.23 & 76.2\\
        Seg & Cityscapes & Cityscapes & 0.63 & 77.4\\ 
    \bottomrule
    \end{tabular} 
    \caption{
    Ablation on architecture search with domain awareness. 
    Det : Object detection. 
    Seg : Semantic segmentation. 
    Target: Target dataset of downstream task where models are finetuned.
    Source: Source dataset for model selection.
    }
    ~\label{tab:5_ablation_on_data_awareness} 
\end{table} 

\section{Limitation}
\label{sec:limitation}
The major limitation of this work lies in the imbalanced training among subnets. 
Specifically, we find that the smaller two-thirds of subnets in supernet are trained well while the rest larger subnets are not. We infer that subnets could benefit more from knowledge distillation and most of their weights are always covered regardless of which subnet is sampled during training. Conversely, larger subnets have no effective teacher as source of knowledge and their exclusive weights might be under insufficient training. 

For our assumption, it indeed can not deal with this situation where candidate subnets are close to the largest teacher network. Because when one model architecture outperforms the largest one, its output features are also far away from the counterpart of largest teacher.

\section{Conclusion}
We have explored combining NAS with self-supervised learning and positive results are shown. Firstly, we manage to train massive weight-sharing subnets in a supernet simultaneously in the regime of self-supervised learning. 
More importantly, this mechanism of supernet training makes possible the unlabeled NAS since the feature distance between subnets and the largest network works perfectly as a self-supervised metric for model selection. 
Our work is orthogonal to most existing self-supervised learning methods and endows them the capability of customization on various downstream needs.
We hope our approach could truly useful in real-world applications and our adventure on NAS in SSL could inspire more genius minds.

\section{Acknowledgement}
We thank Jiawei He, Shuwei Sun, Yuqi Wang, Lin Zhang and anonymous reviewers for their helpful discussions that improved this paper. 

This work was supported in part by 
the Major Project for New Generation of AI (No.2018AAA0100400),
the National Natural Science Foundation of China (No. 61836014, No. U21B2042, No. 62072457, No. 62006231).

{\small
\bibliographystyle{ieee_fullname}
\bibliography{egbib}
}

\end{document}